\documentclass[11pt]{article}

\usepackage{acl}
\usepackage{times}
\usepackage{latexsym}

\usepackage[T1]{fontenc}

\usepackage[utf8]{inputenc}

\usepackage{microtype}

\usepackage[normalem]{ulem}
\useunder{\uline}{\ul}{}

\usepackage{booktabs}

\usepackage{graphicx}

\usepackage{multirow}

\title{HFL at SemEval-2022 Task 8: A Linguistics-inspired Regression Model with Data Augmentation for Multilingual News Similarity}

\author{
Zihang Xu$^\dag$,
Ziqing Yang$^\dag$,
Yiming Cui$^\ddag$$^\dag$,
Zhigang Chen$^\dag$ \\
{$^\dag$State Key Laboratory of Cognitive Intelligence, iFLYTEK Research, China} \\
{$^\ddag$Research Center for SCIR, Harbin Institute of Technology, Harbin, China 
} \\
$^\dag$\tt\{zhxu13,zqyang5,ymcui,zgchen\}@iflytek.com \\
$^\ddag$\tt ymcui@ir.hit.edu.cn}

\begin{document}
\maketitle
\begin{abstract}

This paper describes our system designed for SemEval-2022 Task 8: Multilingual News Article Similarity. We proposed a linguistics-inspired model trained with a few task-specific strategies. The main techniques of our system are: 1) data augmentation, 2) multi-label loss, 3) adapted R-Drop, 4) samples reconstruction with the head-tail combination. We also present a brief analysis of some negative methods like two-tower architecture. Our system ranked 1st on the leaderboard while achieving a Pearson's Correlation Coefficient of 0.818 on the official evaluation set.

\end{abstract}

\section{Introduction}

In Task 8 \cite{taskpaper}, we are expected to assess the similarity of pairs of multilingual news articles. Ten different languages are covered in this task, including Spanish, Italian, German, English, Chinese, Arabic, Polish, French, Turkish and Russian. Task 8 emphasizes more the events themselves described in the news rather than the style of writing or other subjective characteristics. Therefore, it is beneficial to improve the quality of clustering of news articles and to explore similar news coverage across different outlets or regions.

The foundation model \cite{foundationmodel} we choose is XLM-RoBERTa (XLM-R) \cite{xlmr} which has been proved to be a powerful multilingual pre-trained language model compared with other models like mBERT \cite{mbert} and it can process all the languages existing in Task 8. Based on that, a great variety of strategies have been tested along with our exploration like data augmentation (DA), head-tail combination, multi-label loss, adapted R-Drop, etc.

Through this task, we realized the importance of data quality and efficient training schemes in a cross-lingual setting. By struggling to improve the richness of the data and find out what methods are effective when training such a similarity assessment model, our system\footnote{Our code will be available at \url{https://github.com/GeekDream-x/SemEval2022-Task8-TonyX}} ranked 1st in this competition.

\section{Background}

\subsection{Dataset Description} \label{taskdescription}

There are 4,964 samples with 8 language pairs in the training set and the test set contains 4,593 samples in 18 different language pairs, the details of which are presented in Table \ref{dataset}. Due to some inaccessible URLs, the training set is slightly smaller than it should be (22 samples missing in total).

The similarity scores of pairs of articles in provided dataset are rated on a 4-point scale (between 1 and 4) from most to least similar from 7 sub-dimensions, including \textit{Geography}, \textit{Entities}, \textit{Time}, \textit{Narrative}, \textit{Overall}, \textit{Style} and \textit{Tone} (an example is provided in Appendix). However, only the predictions for \textit{Overall} will be used to evaluate the performance of our systems.

\subsection{Related Work}

Research on text similarity always attracts people's eyes as it acts as the basis of quite a few NLP downstream tasks like information retrieval \cite{infomationretrieval}. Previously, some methods based on statistics like BM25 \cite{bm25} and Edit Distance \cite{editdistance} are used to evaluate the relevance between two texts but they do not work anymore in cross-lingual settings. Then, after dense word embedding in low dimensions like Word2Vec \cite{word2vec} was put forward, methods like calculating the cosine similarity \cite{semanticcosinesimilarity} with the sentence embedding based on each word embedding came into use. However, it is hard for these approaches to capture the latent meaning of the whole article precisely. Nowadays, depending on transformer-based general pre-trained models are becoming the new paradigm and plenty of models for multilingual and cross-lingual settings have been proposed like mBERT \cite{mbert}, ERNIE-M \cite{erniem} and XLM-R \cite{xlmr}.

\section{System Overview} \label{systemoverview}

\begin{figure}
    \centering
    \includegraphics[scale=0.38]{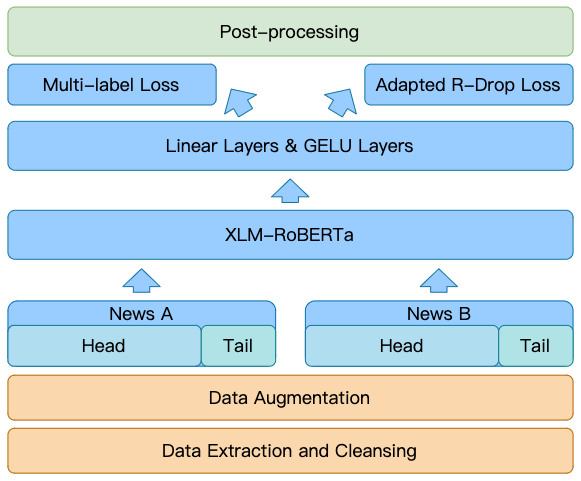}
    \caption{The overall framework of our system proposed for SemEval-2022 Task 8.}
    \label{fig:systemdiagram}
\end{figure}

Our baseline system is simply providing a pair of articles to XLM-R and regressing its output from [CLS] token to the manually annotated similarity score by training with Mean Squared Error (MSE). All the optimization methods discussed below are applied based on this architecture and the overall framework of our final system is illustrated in Figure \ref{fig:systemdiagram}. After training with all the positive strategies, we then made an ensemble of the best models on each fold for the final prediction.

\subsection{Data Augmentation}

In this task, we augmented the training data in two different ways and they will be introduced respectively in the following subsections.

\subsubsection{Back Translation}

It is clear from Table \ref{dataset} that the original training set is not sufficient to train XLM-R, so we made use of back-translation to enrich it. As the English pairs account for the largest, we only paid attention to the non-English samples in this stage. Take the French samples for example, by calling Google Translation API\footnote{\url{https://cloud.google.com/translate}}, we translate the French articles to English and then translate the English texts back to French. As for the cross-lingual pairs with German and English, we only back-translate the German part and then combine it with the corresponding English part to form a new sample.

\begin{table*}
    \small
    \setlength\tabcolsep{2pt}
    \centering
    \begin{tabular}{llllllllllllllllllll}
    \toprule 
                   & \textbf{ar} & \textbf{de} & \textbf{en} & \textbf{es} & \textbf{fr} & \textbf{it} & \textbf{pl} & \textbf{ru} & \textbf{tr}  & \textbf{zh} & \textbf{de-en} & \textbf{de-fr} & \textbf{de-pl} & \textbf{es-en} & \textbf{es-it} & \textbf{fr-pl} & \textbf{pl-en} & \textbf{zh-en} & \textbf{Total} \\ \midrule
                   
    \textbf{Train} & 274         & 857         & 1787        & 567         & 72          & 0           & 349         & 0           & 462 & 0           & 574            & 0              & 0              & 0              & 0              & 0              & 0              & 0              & 4942           \\
    \textbf{Test}  & 298         & 611         & 236         & 243         & 111         & 442         & 224         & 287         & 275 & 769         & 190            & 116            & 35             & 498            & 320            & 11             & 64             & 223            & 4953           \\
    \textbf{Train+DA}    & 548         & 1714        & 1787        & 1134        & 461         & 586         & 689         & 401         & 924 & 800         & 1148           & 317            & 0              & 586            & 586            & 0              & 349            & 800            & 12830 \\      
    
    \bottomrule
    \end{tabular}
    \caption{\label{dataset}
    Data distribution in each set. Columns with one language (e.g. ``zh'') mean the two articles in a pair are in the same language. Columns with two languages (e.g. ``zh-en'') indicate the corresponding cross-lingual pairs.
    }
    \end{table*}

    \begin{table}
    \centering
    \begin{tabular}{ccc}
    \toprule
    \textbf{Origin} &\textbf{Quantity} & \textbf{Target} \\
    \midrule
    \multirow{3}{*}{en-en} & 401 & ru-ru \\
     & 800 & zh-zh / zh-en \\
     & 586 & it-it / es-en / es-it \\
    pl-pl & 349 & pl-en \\
    de-en & 317 & de-fr / fr-fr \\
    \bottomrule
    \end{tabular}
    \caption{\label{translate-train arrangement}Arrangement for the construction of translate-train set.
    }
    \end{table}

\subsubsection{Translate Train}

Another weakness of the original training set is the severe lack of some monolingual language pairs which exist in the test set but not in the training set like Chinese and quite a few cross-lingual language pairs like German to French. To deal with this problem, we planned to generate translate-train data to fill the gap. 

In such semantic comprehension tasks, it is undoubted that the richer semantic information is, the better the model performance will be. Therefore, for maintaining the semantic richness to the largest extent, we made an arrangement for the construction of the translate-train set (details are provided in Table \ref{translate-train arrangement}).

As the average quantity of non-English monolingual samples in the training set is 430, for the sake of balancing the whole dataset, we decided to round it down to 400 and let it be the number of translated samples for Russian (due to some precision issues, it became 401 accidentally). As we may know, Russian and English both belong to Indo-European Family \cite{russian} while Chinese is a member of the Sino-Tibetan Family \cite{chinese}, which indicates that there are quite a lot of common characteristics between the two languages like syntactic structures and lexical analysis methods. So, the most English samples in the original training set would help more in understanding Russian instead of Chinese. Therefore, we decided to generate more Chinese pairs and here we just doubled the number for Russian. Furthermore, the English samples left were all used for generating samples in Italian and Spanish. 

In order to improve the reusability of those samples newly translated already, some work on recombination among different languages pairs was done in this phase. For instance, translating German to English samples to French would let us get German to French samples in the meantime.

\subsection{Head-tail Combination} \label{headtailintroduction}

There is no doubt that different types of texts have different features. As for news, the title tends to be the most informative place in each article since the authors need to use as few concise words as possible to let the readers know what happened in the story. Besides, we believe the head and tail parts of a news article provide much information as well as similar to the introduction and conclusion parts in a research paper. As the XLM-R is capable of processing 512 tokens in each sequence (a pair of articles) at most and the large majority of articles in the training set are much longer than 256 tokens (see Appendix), we tried different truncation strategies to further boost the model performance.

\subsection{Multi-label Loss} \label{multilabellossintro}

As introduced in Background, only the predictions for \textit{Overall} will be used to evaluate, but the other 6 sub-dimensions are also probably helpful for assisting in building a better model. Consequently, we tried to assign various weights for \textit{Overall} when calculating the loss while treating other sub-dimensions equally. For example, if the loss for \textit{Overall} accounts for 40\%, the percentages of the other six sub-dimensions are all 10\% individually.

\subsection{Adapted R-Drop}\label{rdropoverview}

R-Drop is proved to be an effective regularization method based on dropout, by minimizing the KL-divergence of the output distributions of every two sub-models generated via dropout in model training \cite{liang2021rdrop}. To better fit with this regression task, we replaced the KL-divergence loss with MSE loss (adapted R-Drop). Similarly, at each training step, we feed the samples through the forward pass of the network twice. Then, our adapted R-Drop method tries to regularize the model by minimizing the two predicted scores for the same sample, which is:
$$ L_{R}^{i} =  \textup{MSE}(y_{1}^{i}, y_{2}^{i}) $$
where the $ y_{1}^{i} $ means the model output in the first forward pass for the $ i_{th} $ sample. With the basic MSE loss $ L_{B} $ of the two forward passes:
$$ L_{B}^{i} = \frac{1}{2} \cdot (\textup{MSE}(y_{1}^{i}, \hat y^{i}) + \textup{MSE}(y_{2}^{i}, \hat y^{i})) $$
where the $\hat y^{i} $ is the label of the $ i_{th} $ sample, the final training target is minimizing $ L^{i} $ for $ i_{th} $ sample:
$$ L^{i} = \alpha \cdot L_{R}^{i} + (1-\alpha) \cdot L_{B}^{i}$$
where the $ \alpha $ is the weight for the adapted R-Drop loss. Based on this description, it is easy to extend the formulas to the one of forwarding three times.

\subsection{Extra Linear Layers}

In our baseline system, the prediction score is generated by passing the output of [CLS] token from XLM-R through a single linear layer with the size of (1024, 1). In other words, there are only 1024 parameters that are responsible for the regression from the sentence representation vector to the prediction score, which is probably beyond their power. Hence, we attempted to add a few more layers on top of the XLM-R.

\subsection{Post-processing}

Once getting the prediction scores, we further corrected some wrong numbers which were outside the expected range. As introduced in Section \ref{taskdescription}, the annotators annotated the similarity in the range (1, 4); consequently, we clipped the outliers.

\section{Experimental setup}

\subsection{Dataset Split}

Both the original training set and the training set with DA set were split into 10 subsets with no intersection by random sampling. All the experiments discussed in this paper were conducted with 10-fold cross-validation, and the results displayed are the averages. By using the cross-validation method \cite{crossvalidation}, we could ensure the strategies applied will take a good effect on the final test set to the largest extent.

\subsection{Pre-processing}

The news articles in all the data sets are released as URLs and the task organizers offer us a python script\footnote{\url{https://github.com/euagendas/semeval\_8\_2022\_ia\_downloader}} which helps to download the pages. After downloading the original files in JSON format, we then extracted and combined ``title'' and ``text'' parts of each article and abandoned all other information like ``description''. Before starting training our model, apart from conducting data augmentation to the training set, we also cleaned the data and joined the head and tail parts of each article. During the process of cleaning, we mainly removed some dirty formatted data like URLs and file paths.

\subsection{Evaluation Metrics}

The evaluation metric for task 8 is the Pearson's Correlation Coefficient (Pearson's CC) which is a measure of linear correlation between two series of data with a range from -1 to 1 (from least to most correlated) \cite{pearson}.

\subsection{Others}

Although hyper-parameters tuning is not a crucial point in our work, we tested a few values for several of them within a small range as they did have an influence on our decisions about how well a strategy worked (see Appendix). Additionally, to help readers replicate our experiments, the details of tools and libraries are provided (see Appendix).

\begin{table}[t]
    \centering
    
    \begin{tabular}{lc}
    \toprule
    \multicolumn{1}{l}{\textbf{System}} & \multicolumn{1}{l}{\textbf{Pearson's CC}} \\ \midrule
    \multicolumn{2}{c}{\textit{w/ data augmentation}} \\ \midrule
    Baseline & 83.49 \\
    \ \ \ \ + DA & \textbf{85.86} \\ \midrule
    \multicolumn{2}{c}{\textit{w/o data augmentation}} \\ \midrule
    Baseline & 84.94 \\
    \ \ \ \ + Head-tail Combination & 85.38 \\
    \ \ \ \ + Multi-label Loss & 85.33 \\
    \ \ \ \ + Adapted R-Drop & \textbf{86.14} \\
    \ \ \ \ + Extra Linear Layers & 85.50 \\ 
    \bottomrule
    \end{tabular}
    \caption{\label{WholeResults}Best results with training methods we used.}
    \end{table}

\section{Results}

\subsection{Overall Performance}

Finally, our system got 0.818 on the evaluation set according to the official scoring system and ranked 1st. As results are shown in Table \ref{WholeResults}, all the strategies introduced in Section \ref{systemoverview} turned out to have positive effects, and we will discuss the effect of the strategies mentioned individually in the following subsections. For convenience, all the results from our experiments are multiplied by 100.

\subsection{Data Augmentation}

To find out whether the augmented data was helpful or not, we trained our system on the original training set and the training set with DA respectively (samples used for testing were removed in both of them), then tested it on each fold of the DA set. In experiments on other strategies, we trained and tested our system on the original training set. And this is the difference between the two baselines in Table \ref{WholeResults}.

Without any surprise, an evident increase is observed from the results displayed in the top part of Table \ref{WholeResults}, based on which we could make a conclusion that a more abundant training set is definitely beneficial for building a strong system.

\subsection{Head-tail Combination}

As introduced in Section \ref{headtailintroduction}, we realized the importance of the head and tail parts of the news articles. However, we cannot determine which part should be paid more attention to heuristically. So, we tried on different ratios of head-tail combination and the results are enumerated in Table \ref{headtailresults}. Clearly, the head part plays a much more important role by comparing the first and last rows where only either of them are used. However, from the middle three rows where the head and tail parts are combined, it is indicated that the tail part also benefits the whole model performance.

\begin{table}[t]
\centering
\begin{tabular}{llc}
\toprule
\textbf{Head} & \textbf{Tail} & \textbf{Pearson's CC} \\ \midrule
256 & 0 & 84.94 \\
200 & 56 & \textbf{85.38} \\
128 & 128 & 85.21 \\
56 & 200 & 84.53 \\
0 & 256 & 78.85 \\ 
\bottomrule
\end{tabular}
\caption{\label{headtailresults}Results on different head-tail combinations.}
\end{table}

\subsection{Multi-label Loss}

As discussed in Section \ref{multilabellossintro}, we used other 6 dimensions and assigned a few different values for the weight of \textit{Overall} from 0\% to 100\%. It is explicitly observed from Table \ref{multilabelresults} that there is an overwhelming increase followed by a slight drop while the weight of \textit{Overall} rises gradually. Based on the experiment results, we believe that \textit{Overall} is of the greatest importance to this task, yet the other 6 sub-dimensions also have a positive effect on achieving a better similarity assessment system.

\begin{table}[]
\centering
\begin{tabular}{lc}
\toprule
\textbf{\textit{Overall} Weight} & \textbf{Pearson's CC} \\ \midrule
0\% & 45.17 \\
30\% & 85.07 \\
50\% & 85.31 \\
75\% & \textbf{85.33} \\
100\% & 84.94 \\ 
\bottomrule
\end{tabular}
\caption{\label{multilabelresults}Results on training with multi-label loss.}
\end{table}

\subsection{Adapted R-drop}

As described in Section \ref{rdropoverview}, the training loss in our system is composed of both the loss between predictions and labels and the loss between the predictions from different forwarding processes. Here, we explored forwarding once to three times while changing the weight of adapted R-Drop loss.

Apparently, there is a phenomenon from Table \ref{rdropresults} that no matter how large the weight of R-Drop loss is, the more forwarding times are, the better results we will achieve. However, by comparing the results between forwarding once and twice and the results between forwarding twice and three times, we speculate that there is a marginal utility \cite{marginalutility} on this trick, which means the additional benefit from this method will decrease while simply continuing increasing the number of forwarding.

\begin{table}[]
\centering
\small
\begin{tabular}{lccccc}
\toprule
\textbf{RD Weight} & 1\% & 5\% & 10\% & 30\% & 50\% \\ \midrule
\textbf{1F} & 84.94 & 84.94 & 84.94 & 84.94 & 84.94 \\
\textbf{2F} & 85.77 & 85.85 & 85.78 & 86.03 & 85.99 \\
\textbf{3F} & \textbf{85.95} & \textbf{86.14} & \textbf{85.98} & \textbf{86.13} & \textbf{86.07} \\
\bottomrule
\end{tabular}
\caption{\label{rdropresults}
Results about adapted R-Drop (RD) in different settings. ``2F'' means forwarding twice.}
\end{table}

\subsection{Extra Linear Layers}

During the process of exploration in this direction, we attempted to add 2 or 3 extra linear layers to test if it worked. In the 2-layer setting, the sizes of the layers are (1024, 512) and (512, 1) while sizes composed of (1024, 768), (768, 256) and (256, 1) are prepared for the 3-layer setting. Two sets of experiments were conducted in both settings about whether to put an activation layer (we used GELU \cite{gelu} here) between adjacent linear layers or not.

It can be observed from Table \ref{addlinearlayersresults} that there is only a quite small difference that caused by activation layers in each setting and the effect of that is not always positive. In addition, by comparing the results from different settings, we could draw a conclusion that more parameters did help to boost the system performance even if the benefit does not show linear growth.

\begin{table}[]
\centering
\begin{tabular}{lc}
\toprule
\textbf{System} & \textbf{Pearson's CC} \\ \midrule
1-layer & 84.94 \\
2-layer & 85.46 \\
\ \ \ \ + activation & \textbf{85.50} \\
3-layer & 85.32 \\
\ \ \ \ + activation & 85.23 \\ 
\bottomrule
\end{tabular}
\caption{\label{addlinearlayersresults}
Results on different extra layers.}
\end{table}

\subsection{Negative Results}

Aside from the strategies discussed above, several tricks that were attempted to deploy in our system as well turned out to be meaningless or had a bad effect on the model performance. For example, we tried to use a pooling vector (max or mean) or the fusion of [CLS] vectors from different layers in XLM-R as the article representation. We also tried to expand the length of sentences that XLM-R could process to 1024 tokens by modifying its position embedding matrix by means of adding a random shift vector after each vector or just randomly initializing the latter part of the learnable expanded matrix. Each negative strategy mentioned above brought approximately at least 2 points drop on the Pearson's CC. Furthermore, unsurprisingly, a two-tower architecture where each shared-parameter model processed each article in a pair led to scores of points decrease, which reflected the importance of semantic interaction via the attention mechanism inside the model.

\subsection{Error Analysis}

After the evaluation phase ended, the evaluation data with labels were provided and we also checked the system performance on different language pairs individually. The details of our best submission are presented in Table \ref{resultsinbestsubmissiontable}. It is obvious that the model tends to perform worse on the language pairs which are rare or absent from the training set like German to Polish (only 64.31). Interestingly, although having seen monolingual samples in Polish and related cross-lingual data, the system still behaves badly on Polish monolingual data (just slightly over 75), which is probably due to its complicated lexical variation and grammar rules \cite{polish}. 

\begin{table}[]
\centering
\begin{tabular}{cccccc}
\toprule
\textbf{en} & \textbf{de} & \textbf{es} & \textbf{pl} & \textbf{tr} \\ \midrule
87.19 & 84.96 & 86.64 & 75.29 & 83.54 \\ \midrule
\textbf{ar} & \textbf{ru} & \textbf{zh} & \textbf{fr} & \textbf{it} \\ \midrule
79.42 & 78.47 & 76.78 & 86.53 & 86.17 \\ \midrule
\textbf{es-en} & \textbf{de-en} & \textbf{pl-en} & \textbf{zh-en} \\ \midrule
86.35 & 85.98 & 88.18 & 81.00 \\ \midrule
\textbf{es-it} & \textbf{de-fr} & \textbf{de-pl} & \textbf{fr-pl} \\ \midrule
81.97 & 68.89 & 64.31 & 82.68 \\
\bottomrule
\end{tabular}
\caption{\label{resultsinbestsubmissiontable}Individual results of all language pairs in our best submission.}
\end{table}

\section{Conclusion}

By deploying various optimization methods, including data augmentation, head-tail combination, multi-label loss, adapted R-Drop and adding extra linear layers, we built a relatively strong system for assessing the similarity between a pair of news articles in multilingual and cross-lingual settings and ranked 1st in the competition with a Pearson's CC of 0.818 on the official evaluation set.

In the future, apart from enriching the training data, we are also supposed to analyze the languages individually and try to leverage the exclusive rules or features of each language rather than relying too heavily on general pre-trained models to further boost the model performance, especially on those minority languages.

\bibliography{custom}
\bibliographystyle{acl_natbib}

\appendix

\section{Appendix}

Table \ref{hyperparameters} and Table \ref{libraries} provide the details about the corresponding hyper-parameters and libraries respectively, which are beneficial to help replicate our experiments.

\begin{figure*}
\centering
\includegraphics[scale=0.35]{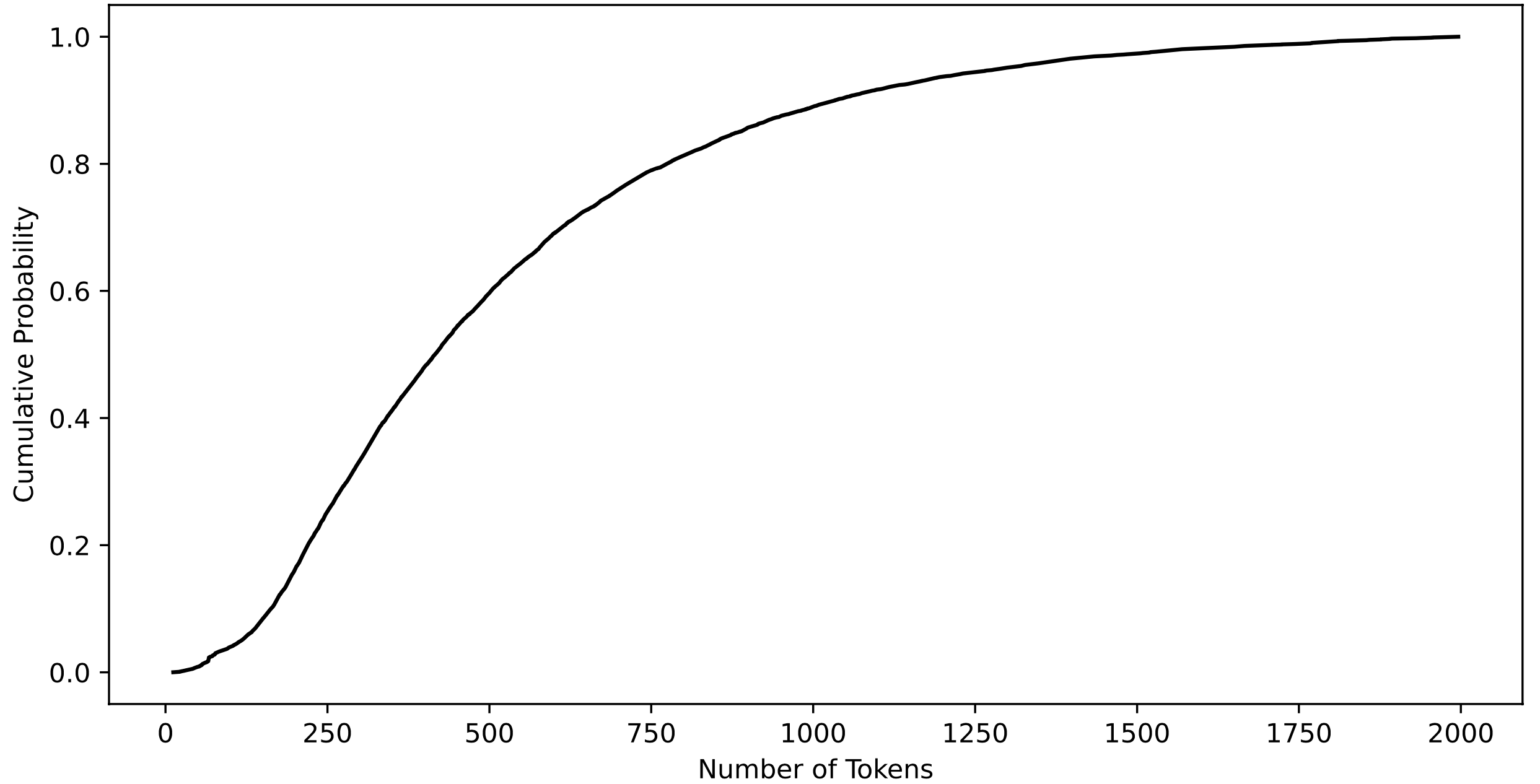}
\caption{Cumulative probability distribution of article lengths in the training set.}
\label{fig:data_length_distribution}
\end{figure*}

Figure \ref{fig:data_length_distribution} illustrates the length distribution of samples in the training set.

Table \ref{sample} gives an example of samples in the training set.

\begin{table}
\centering
\begin{tabular}{lc}
\toprule
\textbf{Hyperparameter} & \textbf{Range/Value} \\
\midrule
Epoch & 20 \textasciitilde\  30 \\
Batch Size & 32 \\
Weight Decay & 1e-4 \\
Warm-up Rate & 0.1 \\
Learning Rate & 5e-6 \textasciitilde\  3e-5 \\
\textit{Overall} Weight & 0 \textasciitilde\  1 \\
Adapted R-Drop Weight & 0.01 \textasciitilde\  0.5 \\
\bottomrule
\end{tabular}
\caption{Main hyper-parameters tuned in our system.}
\label{hyperparameters}
\end{table}

\begin{table}
\centering
\begin{tabular}{lc}
\toprule
\textbf{Tools \& Libraries} & \textbf{Version} \\
\midrule
NumPy & 1.21.2 \\
pandas & 1.2.4 \\
Python & 3.7.10 \\
PyTorch & 1.9.0 \\
Transformers & 4.5.1 \\
semeval\_8\_2022\_ia\_downloader & 0.1.7 \\
\bottomrule
\end{tabular}
\caption{Main tools and libraries used in our system.}
\label{libraries}
\end{table}

\begin{table*}[]
\begin{tabular}{lp{13cm}}
\toprule[1pt]
\textbf{Key} & \textbf{Value} \\ \midrule
\textbf{pair\_id} & 1626170156\_1623571850 \\ 
\textbf{lang1} & de \\
\textbf{lang2} & en \\ \midrule
\textbf{text1} & US-Bürgerrechtler verklagen Trump wegen Polizeieinsatzes. Der Einsatz am Montag sei gesetzwidrig gewesen, da die Demonstranten sich friedlich verhalten hätten, argumentieren die Bürgerrechtsorganisation ACLU und andere Gruppen in ihrer am Donnerstag eingereichten Klage. Die Klage richtete sich auch gegen Justizminister Bill Barr und Verteidigungsminister Mark Esper. Nach dem massiven Polizeieinsatz war Trump zu Fuss zu einer Kirche nahe seines Amtssitzes gegangen. Die St.-Johns-Kirche war am Tag davor bei den Protesten gegen Rassismus und exzessive Polizeigewalt durch ein Feuer beschädigt und mit Graffiti beschmiert worden. Der Präsident liess sich vor dem Gotteshaus mit der Bibel in der erhobenen Hand fotografieren. Trump habe eine ""kriminelle Attacke"" auf Demonstranten geführt, erklärte der ACLU-Vertreter Scott Michelman. Durch dieses Vorgehen würden ""die Fundamente der Verfassungsordnung der Nation erschüttert"". Der Polizeieinsatz beim Weissen Haus hat in den USA viel Kritik ausgelöst. Trumps designierter Herausforderer bei der Wahl im November, Joe Biden, reagierte empört darauf, dass Trump ""für einen Fototermin"" Tränengas und Gummigeschosse auf Demonstranten habe feuern lassen. Justizminister Barr verteidigte jedoch am Donnerstag den Polizeieinsatz. Dieser habe nichts damit zu tun gehabt, dass Trump sich danach zu Fuss zu der Kirche begeben hatte, beteuerte er. In den USA finden seit vergangener Woche landesweite Anti-Rassismus-Proteste statt. Sie waren durch den Tod des Afroamerikaners George Floyd bei einem brutalen Polizeieinsatz in Minneapolis ausgelöst worden. Im Zuge der Proteste kam es immer wieder zu Ausschreitungen. \\
\midrule
\textbf{text2} & Joe Biden Addresses The Nation On Race And Trump's Attacks On Protesters Via the Washington Post: Seeking to console a nation riven by nights of violence with a promise to heal its racial wounds, former vice president Joe Biden on Tuesday will bluntly criticize President Trump’s decision a night earlier to clear protesters from a Washington street so he could pose with a Bible in front of St. John’s Episcopal Church, according to speech excerpts released in advance. “When peaceful protesters are dispersed by the order of the President from the doorstep of the people’s house, the White House — using tear gas and flash grenades — in order to stage a photo op at a noble church, we can be forgiven for believing that the president is more interested in power than in principle,” the presumptive Democratic presidential nominee plans to say, according to the excerpts released by his campaign. “More interested in serving the passions of his base than the needs of the people in his care,” he plans to add. “For that’s what the presidency is: a duty of care — to all of us, not just our voters, not just our donors, but all of us.” The remarks will be delivered at Philadelphia’s City Hall. Philadelphia was also where Barack Obama delivered a heralded speech on race relations more than 12 years ago, titled “A More Perfect Union.” Part of the Biden speech will speak to the nation’s concerns over police brutality, with plans to use the words of George Floyd — “I can’t breathe” — as a mantra. Floyd, an unarmed black man, died after a police officer knelt on his neck in Minneapolis. \\ \midrule
\textbf{Geography} & 1.0 \\
\textbf{Entities} & 2.0 \\
\textbf{Time} & 1.0 \\
\textbf{Narrative} & 2.0 \\
\textbf{Overall} & 4.0 \\
\textbf{Style} & 2.0 \\
\textbf{Tone} & 1.0 \\ 
\bottomrule[1pt]
\end{tabular}
\caption{An example in the training set.}
\label{sample}
\end{table*}

\end{document}